\documentclass[graybox]{svmult}

\usepackage{mathptmx}           
\usepackage{helvet}             
\usepackage{courier}            
\usepackage{type1cm}            
\usepackage{makeidx}            
\usepackage{graphicx}           
\usepackage{multicol}           
\usepackage[bottom]{footmisc}   
\usepackage{blindtext}          
\usepackage[numbers]{natbib}    
\usepackage[hyphens]{url}       
\usepackage{float}              

\makeindex                      

\graphicspath{{./figures/}}     

\begin{document}

\title*{In-Air Imaging Sonar Sensor Network With Real-time Processing Using GPUs}

\author{Wouter Jansen, Dennis Laurijssen, Robin Kerstens, Walter Daems, Jan Steckel}

\institute{Wouter Jansen, Dennis Laurijssen, Robin Kerstens, Walter Daems, Jan Steckel \at University of Antwerp, Faculty of Applied Engineering - CoSys Lab\\
Groenenborgerlaan 171, Antwerp, Belgium, \email{jan.steckel@uantwerpen.be}}

\maketitle
\thispagestyle{empty}
\abstract*{For autonomous navigation and robotic applications, sensing the environment correctly is crucial. Many sensing modalities for this purpose exist. In recent years, one such modality that is being used is in-air imaging sonar. It is ideal in complex environments with rough conditions such as dust or fog. However, like with most sensing modalities, to sense the full environment around the mobile platform, multiple such sensors are needed to capture the full 360-degree range. Currently the processing algorithms used to create this data are insufficient to do so for multiple sensors at a reasonably fast update rate. Furthermore, a flexible and robust framework is needed to easily implement multiple imaging sonar sensors into any setup and serve multiple application types for the data. In this paper we present a sensor network framework designed for this novel sensing modality. Furthermore, an implementation of the processing algorithm on a Graphics Processing Unit is proposed to potentially decrease the computing time to allow for real-time processing of one or more imaging sonar sensors at a sufficiently high update rate.}

\abstract{For autonomous navigation and robotic applications, sensing the environment correctly is crucial. Many sensing modalities for this purpose exist. In recent years, one such modality that is being used is in-air imaging sonar. It is ideal in complex environments with rough conditions such as dust or fog. However, like with most sensing modalities, to sense the full environment around the mobile platform, multiple such sensors are needed to capture the full 360-degree range. Currently the processing algorithms used to create this data are insufficient to do so for multiple sensors at a reasonably fast update rate. Furthermore, a flexible and robust framework is needed to easily implement multiple imaging sonar sensors into any setup and serve multiple application types for the data. In this paper we present a sensor network framework designed for this novel sensing modality. Furthermore, an implementation of the processing algorithm on a Graphics Processing Unit is proposed to potentially decrease the computing time to allow for real-time processing of one or more imaging sonar sensors at a sufficiently high update rate.}

\section{Introduction}
\label{sec:introduction}
Sonar sensing has, contrary to what nature tells us, often been judged to as not being capable of supporting complex, dynamic environments for intelligent interaction when used outside water. For implementation in robotics, optical sensors are often the first choice \cite{Brooks1986ARobot} and sonar is not considered. However, as we can see with several species of bats, their biosonar allows for classification of objects and finding their prey among them \cite{Schnitzler2003FromBats, Griffin1974ListeningMen}. In recent years, our research group has developed sonar systems inspired by the echolocation system used by these bats, capable of finding detailed 3D information from complex environments \cite{Steckel2019RTIS, Kerstens2019ERTIS:Applications}. \\
This sensing modality can serve a wide range of applications, especially in harsh environments where optical techniques tend to fail due to medium distortions due to foggy weather or dusty environments. As an example application in robotics, a Simultaneous Localisation and Mapping (SLAM) solution was developed using 3D sonar for estimating the ego-motion of a mobile platform \cite{Steckel2013BatSLAM:Sonar}. Furthermore, in 2017, a control strategy was developed to navigate a mobile platform through an environment using the acoustic flow field generated by the sonar sensor \cite{Steckel2017AcousticSensor}. As these works show, sonar is an increasingly capable sensing modality.\\
However, often multiple sensors are embedded into a (mobile) platform, sensing in different directions. This is no different for sonar. The current sonar system we use is capable of sensing the entire frontal hemisphere in front of the sensor. When trying to sense the entire 360-degree around a mobile platform, multiple sonar sensors are required. In this paper we propose a framework to deploy a network of these sonar sensors for such a platform, providing time-synchronised measurements which can be processed in real-time on a Compute Unified Device Architecture (CUDA)\cite{NVIDIA2019CUDAToolkit} enabled Graphics Processing Unit (GPU) by NVIDIA. Our main focus lies on being able to scale up of the amount of sensors, general ease of use of the framework and being able to always operate in real-time, being only limited by the frequency of sonar measurements.

\section{Sonar Synchronisation}
\label{sec:sonar}
\begin{figure}
\includegraphics[width=\linewidth]{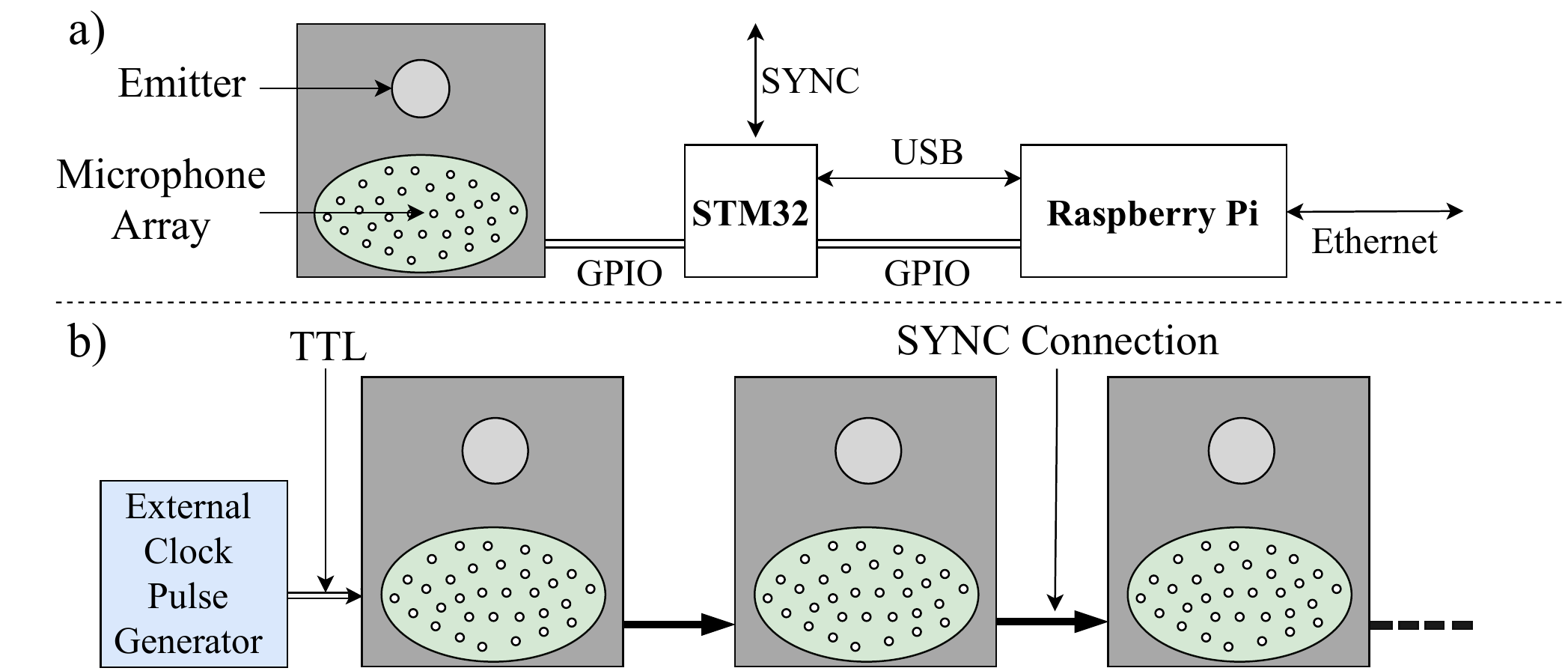}
\caption{a) Schematic overview of all elements of an eRTIS sensor connected to a Raspberry Pi. b) Chain of eRTIS sensors connected with the SYNC connection and initiated by a external pulse generator.}
\label{fig:ertisnetwork}
\end{figure}

First, we will provide a brief overview of the 3D sonar system used, our Embedded Real Time Imaging Sonar (eRTIS) \cite{Steckel2019RTIS}. A much more in-depth description of the sonar can be found in \cite{Kerstens2019ERTIS:Applications, Steckel2013BroadbandNavigation}. The sensor has a single emitter and uses a known, irregularly positioned array of 32 digital microphones, as can be seen on Figure \ref{fig:ertisnetwork}a. \\
A broadband FM-sweep is emitted allowing the microphones to capture the reflected echos. The processing of the recorded microphone data to create the 2D or 3D images of the environment which will be further discussed in Subsection \ref{sec:sonarprocessing}. The microphone signals are recorded over a 1 bit Sigma-Delta Pulse-Density Modulation (PDM) ADC. The embedded microprocessor on the eRTIS sensor sends out the recorded data over a USB 2.0 connection to a Raspberry Pi, which will distribute it to the network which be discussed in the next section. \\
As we use acoustic echolocation, allowing each eRTIS sensor to freely choose when to emit the FM-sweep and capture data with the microphones would cause heavy interference from one another as the emission signals from other sensors would be captured. This would cause artefacts making most measurements meaningless. Therefore, we have to make the emitters of the eRTIS sensors send their FM-sweeps simultaneously in order to make the emitters form a consistent source. To that end, we use a synchronisation mechanism that we have built into the hardware, which is discussed in more detail in \cite{Kerstens20193DNetwork}. The eRTIS sensors use a SYNC signal to synchronise with each other as seen on Figure \ref{fig:ertisnetwork}b. This SYNC signal allows each eRTIS sensor to synchronise with the previous eRTIS sensor and will also propagate the SYNC signal to further daisy-chained eRTIS sensors in the network. To set the measurement rate, the first eRTIS sensor in the chain can be controlled by an external clock pulse. Each connected Raspberry Pi will receive a specific message over its USB serial data connection after the FM-sweep has been emitted and will cause the Raspberry Pi to initiate the capture of the measurement data over that same serial connection. The measurement delay with this synchronised system is constant and is ca. 400 ns. However, this is inconsequential as this is but a small fraction of measurement cycle of an eRTIS sensor sampling at 450 kHz using a update rate of 20 Hz.

\section{Network}
\label{sec:network}
\begin{figure}
\includegraphics[width=\linewidth]{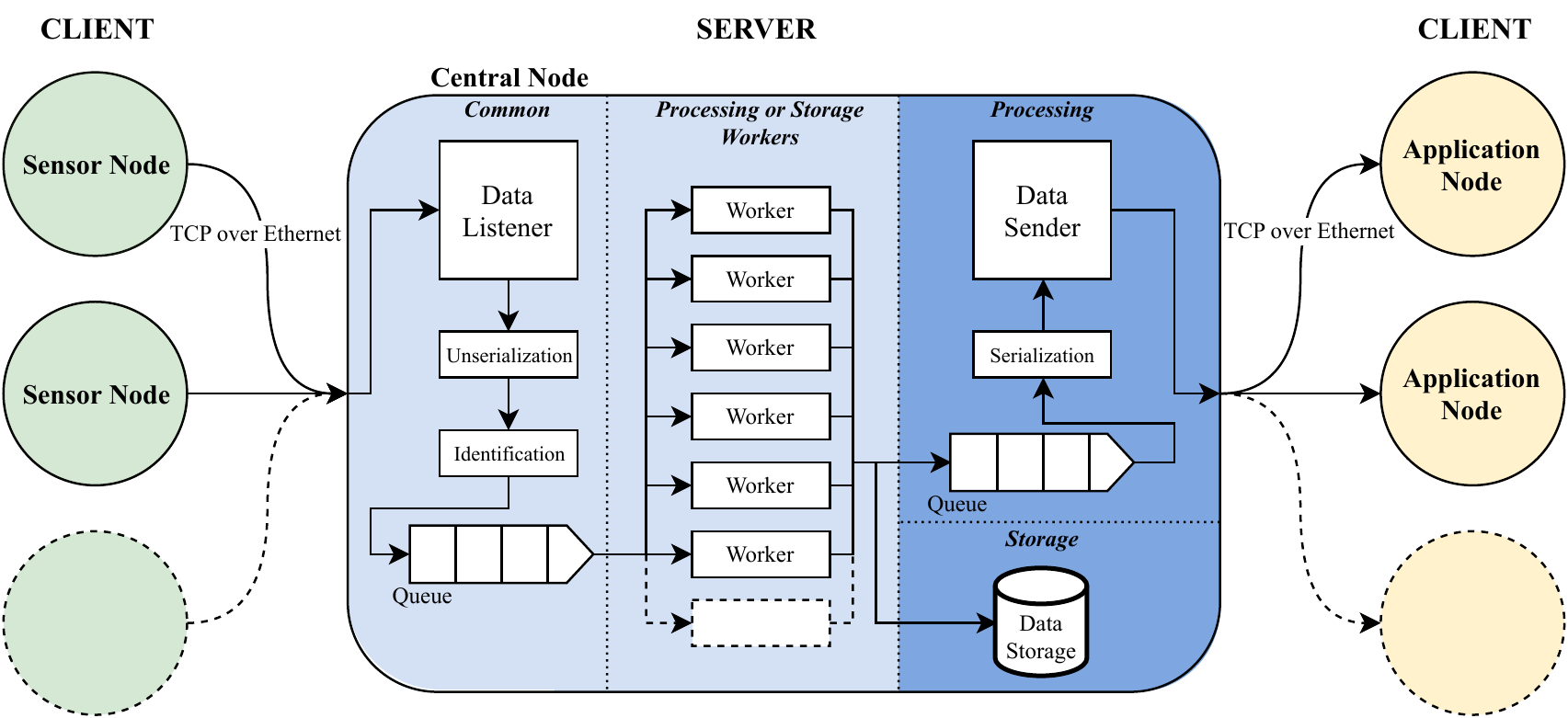}
\caption{A diagram visualising the network topology and communication between all the nodes as well as the different states of the central node. The sensor nodes (client) send their data packages to the central node (server). This node, being in either storage or processing state, will identify each package and add it to a queue. The centre node has a series of multi-threaded workers which will pick from that queue and either store or process the measurement. In the processing state, the resulting
processed imaging data will be put on another queue. Data in this queue is send to the correct
application nodes (client) by separate process.}
\label{fig:communication}
\end{figure}

The chain of hardware-synchronised sonar sensors will continuously capture the recorded data of the microphone array. However, we need additional nodes in this network to receive these data streams and process or store them elsewhere. Furthermore, several applications need to be able to make use of the processed sensor data. This network with various node types needs a particular topology as well as a communication pipeline that is capable of working in real-time. In this section we will discuss these further. For the network topology we use a star format as seen on Figure \ref{fig:communication}. The Raspberry Pi of each eRTIS sensor is connected to the central node.  \\
This central node represents the receiver of all sonar measurements. It is currently configurable in two states. The first state is to store all raw measurement data. It will identify, index and store the recorded data received from each eRTIS sensor. The second state is when real-time processing is required.
The application nodes, the end-users of the processed data, are also directly connected to the central node. \\
This can be various application types such as a navigation safety system, a SLAM algorithm, an acoustic flow control system or a visualisation application. All connections between the nodes are provided by a LAN network on the platform, avoiding any wireless communication, ensuring a fast and reliable Ethernet connection.

\noindent There are a vast amount of proprietary and open-source communication stacks for robotic sensor networks as for example the very commonly used Robot Operating System (ROS) \cite{WillowGarage2019ROS.org}, or a recently developed API called Distributed Uniform Streaming Framework (DUST) \cite{Vanneste2018DistributedProcessing}. However, our imaging system requires a high computational load and therefore is less suited to implement in such existing frameworks directly. Furthermore, the simplicity of the network topology, using a reliable and stable wired LAN network all indicate that a more lightweight, custom-designed solution was a better option. We developed the current version in Python. Nevertheless, we don't exclude the idea of implementing our sensor network into existing frameworks at a later date when our system has more matured. \\
Communication between the various edge nodes and centre node was developed as a uncomplicated client-server communication over TCP. The sensor nodes package and serialise the PDM recorded data combined with an identification serial-number and a timestamp. Afterwards, they send this data package over TCP to the central node. This node, being in either storage or processing state, has a separate server process constantly listening for incoming packages of sensor nodes. Each package is then unserialised, identified and added to a queue. The centre node has a series of multi-threaded processes defined as workers which will pick from that queue and either store or process the measurement. \\
When the centre node is in the processing state, the resulting processed imaging data will be put on another queue. This queue will be picked from by another process, serialising the data and sending it to the correct application nodes. The entire communication layer is shown in Figure \ref{fig:communication}. 

\section{Real-time Processing}
\label{sec:processing}
Where our research group had previously implemented the processing algorithm discussed in this section on a Central Processing Unit (CPU) using Python, it was not capable of performing real-time computation for one or more eRTIS sensors when the amount of directions of interest was increased substantially. To achieve real-time applications making use of our 3D sonar images, an accelerated implementation of the processing pipeline had to be made. First, we will discuss the current processing implementation as it was created for a CPU, followed by the benefits of using a GPU over a CPU and describing its architecture. Finally, we will discuss the changes made to the processing algorithm to make better usage of the GPU resources.

\begin{figure}[H]
\includegraphics[width=\linewidth]{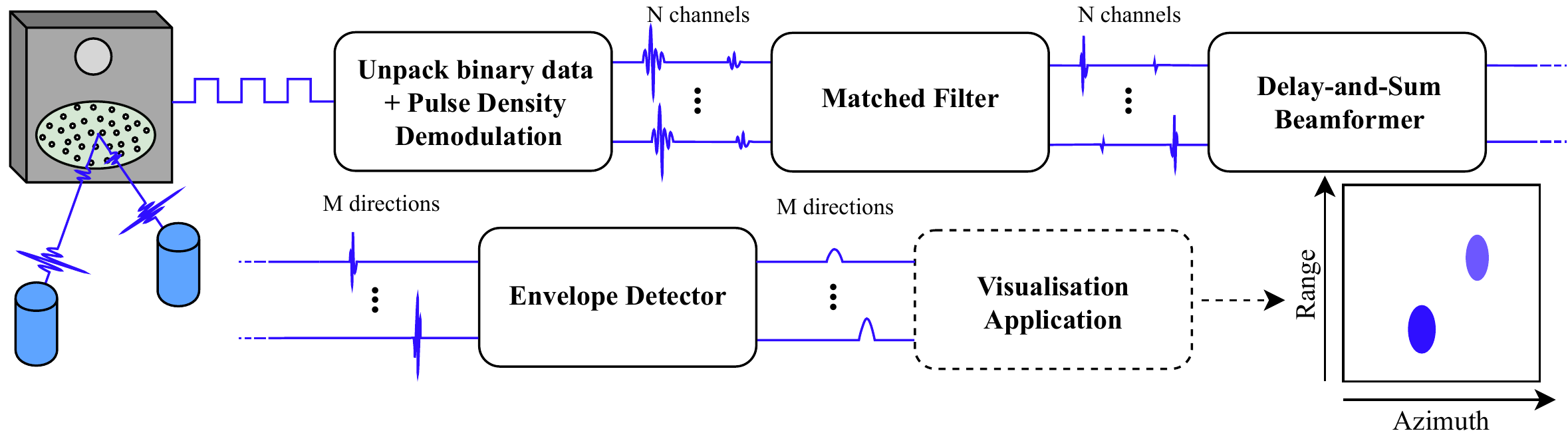}
\caption{Overview of the processing pipeline. The microphone signals containing the reflected echoes of the detected objects are demodulated and passed through a matched filter. Afterwards, a beamformer generates a spacial filter for every direction of interest. The output is then put through an envelope detector for every direction. The last visualisation step and resulting 2D image is not part of the processing but is shown here to illustrate a possible application of the sensor imaging. } 
\label{fig:processingpipeline}
\end{figure}

\subsection{Signal Processing}
\label{sec:sonarprocessing}
For a detailed description on the process that is used to achieve the 3D data with the eRTIS sensor, we advise the reader to read \cite{Steckel2013BroadbandNavigation,Kerstens2019ERTIS:Applications}. A brief overview of the processing pipeline is shown in Figure \ref{fig:processingpipeline}. As initially explained in Section \ref{sec:sonar}, a broadband FM-sweep is emitted and the reflected echoes are recorded by the 32 microphones on the eRTIS sensor. These signals are modulated using PDM and communicated to the processing node over the communication channels discussed in the previous Section. There, after being first demodulated, the signals are processed using a matched filter. \\
Using broadband chirps, we can detect multiple closely spaced reflectors in the reflection signal \cite{Steckel2013BroadbandNavigation}. After this, beamforming takes place. For each direction of interest we can calculate the delay, relative to the particular channel of the microphone array. This allows for steering into the directions of interest with simple delay-and-sum beamforming  \cite{VanTrees2002OptimumProcessing}. After beamforming, we can extract the envelope of the signal as the representation of the reflector distribution for each particular direction, with the time axis representing the range for that direction. We now have a 2D or 3D image that gives us the intensity, range and direction for each detected reflector.

\subsection{GPU Architecture}
\label{sec:gpuarchitecture}
CUDA \cite{NVIDIA2019CUDAToolkit} has enabled programming on the GPU beyond rendering of graphics. This allows to speed up and parallelise heavy computational tasks. Recently, multiple works have been published using GPUs for audio signal processing \cite{Belloch2014Multi-channelGPU, Belloch2013Headphone-BasedAccelerator, MartinSchneider2012TheCancellation} and (sonar) beamforming algorithms such as (Linearly Constrained) Minimum Variance, Minimum Variance Distortionless Response, Generalised Sidelobe Canceller and Capon \cite{Lorente2011ParallelApplications, Asen2014ImplementingImaging,  Buskenes2015AnImaging, Buskenes2013AdaptingSystems}. Similarly, our research group looked at GPUs to provide the potential to perform real-time processing of an eRTIS sonar with a high number of directions during beamforming to create a high resolution 3D image of the sensor's frontal hemisphere. Our processing algorithm, which is computationally expensive and needs to be used in parallel to allow for multiple sensors to be processed simultaneously is therefore an ideal case for using a GPU.\\ The architecture of an NVIDIA GPU has been consistent over the last few generations. It is composed of multiple \textit{Streaming Multiprocessors} (SM). Each SM contains a set of pipelined CUDA cores. For high-speed, high bandwidth memory access, an NVIDIA GPU has access to device memory (\textit{global}) and faster on-chip (\textit{shared}) memory. When programming device code that has to run on the GPU cores, it is also called a \textit{kernel} function. One such kernel can be made to be run concurrently by multiple cores. It is then called a \textit{thread}. Threads are grouped in \textit{thread blocks}, with a maximum of 1024 threads for each block. A block can be designed to organise threads in up to three dimensions. All threads in a block have access to the same shared memory. The GPU will schedule the blocks over each SM. An SM can run up to 32 parallel threads, which is called a \textit{warp}. Using all 32 threads in a warp is important, as not optimally using a single SM will increase latency. During the validation of our system we used 3 different GPUs. The Jetson TX2, GTX 1050 and GTX 1080 Ti representing what could be used on a mobile platform with low power usage, a laptop GPU and a high-end desktop GPU respectively. Their specifications can be found in Table \ref{table:gpuspecs}. All share the same restrictions on thread block dimensions and maximum amount of blocks for each SM. The specific architecture of an NVIDIA GPU requires careful adaptation of the algorithms to exploit it's full power. The changes are detailed in the next subsection.

\begin{table}
\caption{Validation NVIDIA GPU Specifications}
\label{table:gpuspecs}      
\begin{tabular}{p{4.1cm}p{2.4cm}p{2.4cm}p{2.4cm}}
\hline\noalign{\smallskip}
 & Jetson TX2 & GTX 1050 & GTX 1080 Ti  \\
\noalign{\smallskip}\svhline\noalign{\smallskip}
Device memory & 8 GB LPDDR4 & 4 GB GDDR5 & 11 GB GDDR5X\\
CUDA cores & 256 & 640 & 3584\\
Streaming Multiprocessors & 2 & 5 & 28\\
GPU clock speed & 1301 MHz & 1190 MHz & 1582 MHz\\
Memory clock speed & 1600 MHz & 3504 MHz & 5505 MHz\\
Power usage under heavy load & $\pm$15 W & $\pm$50 W & $\pm$250 W\\
\noalign{\smallskip}\hline\noalign{\smallskip}
\end{tabular}
\end{table}

\subsection{GPU Algorithm Implementation}
\label{sec:gpuimplementation}
For this initial implementation of the processing pipeline discussed in Subsection \ref{sec:sonarprocessing}, we mostly looked at a direct application of the algorithm developed for the CPU and rewrite it for the GPU. This CPU algorithm was created in Python, making substantial usage of the NumPy and SciPy packages known for their powerful N-dimensional array and signal processing functions respectively. For this paper we will focus on the aspects that were done differently for optimal usage of the GPU during the rewrite of the algorithm.\\
A re-occurring element seen in many parts of the processing algorithm are the several digital signal filters being used. They are used during the PDM demodulation, the matched filter and envelope detection. In the CPU implementation, we design these filters as Butterworth filters and make usage of SciPy's \texttt{lfilter} function. For performing this faster and paralleled on the GPU, we make use of FFT convolution filters. For this we can make use of the CuFFT CUDA library for paralleled FFT and iFFT execution. Furthermore, we can implement parts of the filter such as the element-wise multiplication of the complex FFTs or removing the group delay introduced by the filter easily using GPU kernels.\\
Outside of the filters, other aspects such as the unpacking of the binary data before the PDM demodulation, subsampling before the matched filter and after envelope detection to increase computational performance and delay-and-sum beamforming were implemented as device kernels. By optimally dividing the dimensions of the matrices used in the algorithm in the correct block sizes to fill a complete warp on an SM, one can make excellent usage of the GPU's resources. As discussed in Subsection \ref{sec:gpuarchitecture}, the used GPUs for this paper share the same limitations for the maximum block dimensions, blocks per SM and threads per blocks. This makes it easier to decide the optimal parameters for the kernels. Moreover, an additional performance increase we achieved was by preallocating device memory before we start the processing node. We noticed that allocating and freeing memory for each worker for each measurement decreased performance substantially. Given that our matrix sizes and input variables are consistent, we can perform this step once and free the device memory when we stop the entire processing node at the end.

\section{Experimental Validation}
\label{sec:result}
For validating the processing on a GPU we used three different models as shown in Table \ref{table:gpuspecs}. For comparison, we used three Intel CPUs. An i7-7567U (2 cores, up to 4 GHz), an i7-7700HQ (4 cores, up to 3.8 GHz) and an i9-7960X (16 cores, up to 4.2 GHz). Similarly as with the GPU choices, these represent what could be expected on a mobile platform with low power usage, a laptop and a high-end desktop respectively. We ran 3 different processing configurations for each hardware setup. First, a 2D image of the horizontal plane in front of the sensor, using 90 directions over the entire azimuth range. A second configuration was a 3D image with 1850 directions distributed between -45 and 45 degrees for both azimuth and elevation. Finally, a 3D image of the full hemisphere in front of the sensor, using 3000 directions. For every configuration, 100 measurements were processed and the computation duration was recorded. The results can be found in Table \ref{table:processingresults}.

\begin{table}
\caption{Computing Time Results - Mean and Standard Deviation}
\label{table:processingresults}      
\begin{tabular}{p{2.8cm}p{2.4cm}p{3.2cm}p{3.1cm}}
\hline\noalign{\smallskip}
& 90 directions & 1850 directions & 3000 directions  \\
\noalign{\smallskip}\svhline\noalign{\smallskip}
CPU i7-7567U & 132.53 (4.89) & 976.33 (14.54) & 1541.41 (20.30)\\
CPU i7-7700HQ & 187.85 (19.94) & 1410.85 (183.50) & 2 080.59 (232.22)\\
CPU i9-7960X & 167.13 (16.88) & 1087.04 (109.72) & 1653.45 (167.03)\\
GPU Jetson TX2 & 68.40 (13.40)  & 585.24 (61.02) & 911.47 (91.97)\\
GPU GTX 1050 & 23.03 (2.38) & 202.37 (20.85) & 323.83 (33.18)\\
GPU GTX 1080 Ti & 5.93 (0.66) & 51.58 (6.38) & 77.02 (7.81)\\
\noalign{\smallskip}\hline\noalign{\smallskip}
\end{tabular}
Standard deviation is listed between brackets. All values are in milliseconds.
\end{table}

\noindent One comment we would like to make when looking at the results is the remarkable performance of the i7-7567U compared to the other CPUs. A partial explanation is that the CPU implementation of the processing algorithm was done in Python, making heavy usage of the NumPy and SciPy packages. Most of their functions are not multi-threaded and the Python Global Interpreter (GIL) limits the Python interpreter to only have one thread in a state of execution at any point in time\cite{Python2019GlobalInterpreterLock}. This severely limits our CPU algorithm to make full usage of the i9-7960X's 16 cores. However, if we want would process multiple eRTIS sensors at the same time, we can make usage of Python's Multiprocessing package to run multiple workers simultaneously, making usage of multiple cores (which we do as explained in Subsection \ref{sec:network} when describing the workers). In such cases, the i9 would outclass the other CPUs in our tests. But for these results we are only looking at single sensor processing. Nevertheless, we did not find an explanation why the i7-7567U performs better in single thread performance than the i9-7960X as on paper the second should be better. Lastly, given these results and after doing additional experiments we found that on the Jetson TX2, suited well for a mobile platform, up to three eRTIS sensors could be processed at 5 Hz when using 90 directions of interest. Moreover, on the GTX 1080 Ti, a single eRTIS sensor could be processed at 10Hz with 3000 directions covering the entire frontal hemisphere.

\section{Conclusion and Future Work}
\label{sec:conclusion}
With the sensor network we built, we can now develop new application types much easier with this framework as a basis. From a mobile platform to a larger desktop system, the network can be easily installed due to its flexibility. Furthermore, our results validate that the initial implementation of our processing algorithm on a GPU already heavily outperforms any CPU tested. Therefore, using a GPU and a processing configuration appropriate for the application and available GPU resources allows for one or more eRTIS sensors to be processed in real-time. Additional research has to be done to optimise the GPU algorithm further, notably for the computationally expensive steps such as beamforming and filters. Furthermore, a comparison should be made to a CPU implementation where instead of Python, an optimised C(++) algorithm is developed. \\
We intend to continue our work on this sensor network framework and GPU/CPU algorithms to support more application use-cases and lower the cost in terms of price and in terms of the computing requirements for all configurations.

%

\bibliographystyle{spbasic} 
\interlinepenalty=10000
\bibliography{paper.bbl}

\begin{thebibliography}{21}
\providecommand{\natexlab}[1]{#1}
\providecommand{\url}[1]{{#1}}
\providecommand{\urlprefix}{URL }
\expandafter\ifx\csname urlstyle\endcsname\relax
  \providecommand{\doi}[1]{DOI~\discretionary{}{}{}#1}\else
  \providecommand{\doi}{DOI~\discretionary{}{}{}\begingroup
  \urlstyle{rm}\Url}\fi
\providecommand{\eprint}[2][]{\url{#2}}

\bibitem[{Asen et~al(2014)Asen, Buskenes, Nilsen, Austeng, and
  Holm}]{Asen2014ImplementingImaging}
Asen JP, Buskenes JI, Nilsen CIC, Austeng A, Holm S (2014) {Implementing capon
  beamforming on a GPU for real-time cardiac ultrasound imaging}. IEEE
  Transactions on Ultrasonics, Ferroelectrics, and Frequency Control
  61(1):76--85, \doi{10.1109/TUFFC.2014.6689777}

\bibitem[{Belloch et~al(2013)Belloch, Ferrer, Gonzalez, Martinez-Zaldivar, and
  Vidal}]{Belloch2013Headphone-BasedAccelerator}
Belloch JA, Ferrer M, Gonzalez A, Martinez-Zaldivar F, Vidal AM (2013)
  {Headphone-Based Virtual Spatialization of Sound with a GPU Accelerator}.
  Journal of the Audio Engineering Society 61(7/8):546--561

\bibitem[{Belloch et~al(2014)Belloch, Bank, Savioja, Gonzalez, and
  Valimaki}]{Belloch2014Multi-channelGPU}
Belloch JA, Bank B, Savioja L, Gonzalez A, Valimaki V (2014) {Multi-channel IIR
  filtering of audio signals using a GPU}. In: 2014 IEEE International
  Conference on Acoustics, Speech and Signal Processing, IEEE, pp 6692--6696,
  \doi{10.1109/ICASSP.2014.6854895}

\bibitem[{Brooks(1986)}]{Brooks1986ARobot}
Brooks R (1986) {A robust layered control system for a mobile robot}. IEEE
  Journal on Robotics and Automation 2(1):14--23,
  \doi{10.1109/JRA.1986.1087032}

\bibitem[{Buskenes et~al(2013)Buskenes, {\AA}sen, Nilsen, and
  Austeng}]{Buskenes2013AdaptingSystems}
Buskenes JI, {\AA}sen JP, Nilsen CIC, Austeng A (2013) {Adapting the minimum
  variance beamformer to a graphics processing unit for active sonar imaging
  systems}. The Journal of the Acoustical Society of America 133(5):3613--3613,
  \doi{10.1121/1.4806739}

\bibitem[{Buskenes et~al(2015)Buskenes, Asen, Nilsen, and
  Austeng}]{Buskenes2015AnImaging}
Buskenes JI, Asen JP, Nilsen CIC, Austeng A (2015) {An Optimized GPU
  Implementation of the MVDR Beamformer for Active Sonar Imaging}. IEEE Journal
  of Oceanic Engineering 40(2):441--451, \doi{10.1109/JOE.2014.2320631}

\bibitem[{Griffin(1974)}]{Griffin1974ListeningMen}
Griffin DR (1974) {Listening in the dark; the acoustic orientation of bats and
  men,}. Dover Publications

\bibitem[{Kerstens et~al(2019{\natexlab{a}})Kerstens, Laurijssen, Schouten, and
  Steckel}]{Kerstens20193DNetwork}
Kerstens R, Laurijssen D, Schouten G, Steckel J (2019{\natexlab{a}}) {3D Point
  Cloud Data Acquisition Using a Synchronized In-Air Imaging Sonar Sensor
  Network}. In: IEEE/RSJ International Conference on Intelligent Robots and
  Systems, To be published

\bibitem[{Kerstens et~al(2019{\natexlab{b}})Kerstens, Laurijssen, and
  Steckel}]{Kerstens2019ERTIS:Applications}
Kerstens R, Laurijssen D, Steckel J (2019{\natexlab{b}}) {eRTIS: A Fully
  Embedded Real Time 3D Imaging Sonar Sensor for Robotic Applications}. In:
  IEEE International Conference on Robotics and Automation, To be published

\bibitem[{Lorente et~al(2011)Lorente, Vidal, Pinero, and
  Belloch}]{Lorente2011ParallelApplications}
Lorente J, Vidal AM, Pinero G, Belloch JA (2011) {Parallel Implementations of
  Beamforming Design and Filtering for Microphone Array Applications}. In:
  Signal Processing Conference, European

\bibitem[{Martin~Schneider et~al(2012)Martin~Schneider, Schuh, and
  Kellermann}]{MartinSchneider2012TheCancellation}
Martin~Schneider M, Schuh F, Kellermann W (2012) {The Generalized
  Frequency-Domain Adaptive Filtering Algorithm Implemented on a GPU for
  Large-Scale Multichannel Acoustic Echo Cancellation}. In: Speech
  Communication; 10. ITG Symposium, VDE Verlag GmbH, p 296

\bibitem[{{NVIDIA}(2019)}]{NVIDIA2019CUDAToolkit}
{NVIDIA} (2019) {CUDA Toolkit}.
  \urlprefix\url{https://docs.nvidia.com/cuda/index.html}

\bibitem[{{Python}(2019)}]{Python2019GlobalInterpreterLock}
{Python} (2019) {GlobalInterpreterLock}.
  \urlprefix\url{https://wiki.python.org/moin/GlobalInterpreterLock}

\bibitem[{Schnitzler et~al(2003)Schnitzler, Moss, and
  Denzinger}]{Schnitzler2003FromBats}
Schnitzler HU, Moss CF, Denzinger A (2003) {From spatial orientation to food
  acquisition in echolocating bats}. Trends in Ecology {\&} Evolution
  18(8):386--394, \doi{10.1016/S0169-5347(03)00185-X}

\bibitem[{Steckel(2019)}]{Steckel2019RTIS}
Steckel J (2019) {RTIS}. \urlprefix\url{https://www.3dsonar.eu/}

\bibitem[{Steckel and Peremans(2013)}]{Steckel2013BatSLAM:Sonar}
Steckel J, Peremans H (2013) {BatSLAM: Simultaneous Localization and Mapping
  Using Biomimetic Sonar}. PLoS ONE 8(1):e54,076,
  \doi{10.1371/journal.pone.0054076}

\bibitem[{Steckel and Peremans(2017)}]{Steckel2017AcousticSensor}
Steckel J, Peremans H (2017) {Acoustic Flow-Based Control of a Mobile Platform
  Using a 3D Sonar Sensor}. IEEE Sensors Journal 17(10):3131--3141,
  \doi{10.1109/JSEN.2017.2688476}

\bibitem[{Steckel et~al(2013)Steckel, Boen, and
  Peremans}]{Steckel2013BroadbandNavigation}
Steckel J, Boen A, Peremans H (2013) {Broadband 3-D Sonar System Using a Sparse
  Array for Indoor Navigation}. IEEE Transactions on Robotics 29(1):161--171,
  \doi{10.1109/TRO.2012.2221313}

\bibitem[{Van~Trees(2002)}]{VanTrees2002OptimumProcessing}
Van~Trees HL (2002) {Optimum Array Processing}. John Wiley {\&} Sons, Inc., New
  York, USA, \doi{10.1002/0471221104}

\bibitem[{Vanneste et~al(2018)Vanneste, de~Hoog, Huybrechts, Bosmans, Sharif,
  Mercelis, and Hellinckx}]{Vanneste2018DistributedProcessing}
Vanneste S, de~Hoog J, Huybrechts T, Bosmans S, Sharif M, Mercelis S, Hellinckx
  P (2018) {Distributed Uniform Streaming Framework: Towards an Elastic Fog
  Computing Platform for Event Stream Processing}. In: International Conference
  on P2P, Parallel, Grid, Cloud and Internet Computing, pp 426--436,
  \doi{10.1007/978-3-030-02607-339}

\bibitem[{{Willow Garage} and {Stanford Artificial Intelligence
  Laboratory}(2019)}]{WillowGarage2019ROS.org}
{Willow Garage}, {Stanford Artificial Intelligence Laboratory} (2019)
  {ROS.org}. \urlprefix\url{http://www.ros.org/}

\end{thebibliography}

\end{document}